\documentclass[review,1p,authoryear]{elsarticle}

\usepackage{dcolumn}% Align table columns on decimal point
\usepackage{bm}% bold math
\usepackage{subfigure}

\usepackage{soul}

\usepackage{amsmath,amssymb,amsfonts}
\DeclareMathOperator*{\argmax}{arg\,max}

\usepackage{graphics}
\usepackage{color}
\usepackage{xcolor}
\definecolor{rev1}{rgb}{0,0,0}

\usepackage{graphicx}% Include figure files
\usepackage{subfigure}
\usepackage{algorithm}
\usepackage{algpseudocode}
\algnewcommand\server{\item[\textbf{Server execution:}]}%
\algnewcommand\client{\item[\textbf{ClientUpdate($k,w$):}]}%

\usepackage{enumitem}
\setlist[itemize]{leftmargin=*}
\setlist[enumerate]{leftmargin=*}

\usepackage{comment}
\usepackage{lipsum}
\usepackage{tabularx}
\usepackage{multirow}
\usepackage{mathrsfs}
\usepackage{stackengine}
\usepackage{accents}

\usepackage{url,microtype}
\usepackage[onehalfspacing]{setspace}
\usepackage[thinc]{esdiff} % For nice derivative commands.
\usepackage{amssymb}
\usepackage{amsmath}
\usepackage{bm}
\usepackage{xcolor}
\usepackage{setspace}
\usepackage{geometry}
\usepackage{tikz}
\usepackage{booktabs} % For \toprule \midrule and \bottomrule commands
\definecolor{green}{rgb}{0,0.6,0}
\usepackage{siunitx}
\usepackage[switch]{lineno}

\usepackage{placeins}

\journal{Elsevier}
\begin{document}
%\linenumbers

\begin{frontmatter}
%\title{Deep neural network enabled corrective source term approach applied to 2D heat diffusion modelling} 
%\title{Federated machine learning and decentralized digital twins of complex dynamical systems} 
\title{Decentralized digital twins of complex dynamical systems} 

\author[omersaddress]{Omer San\corref{mycorrespondingauthor}}
\cortext[mycorrespondingauthor]{Omer San}
\ead{osan@okstate.edu}

\author[omersaddress]{Suraj Pawar}
\ead{supawar@okstate.edu}

\author[adilsaddress]{Adil Rasheed}
\ead{adil.rasheed@ntnu.no}

\address[omersaddress]{School of Mechanical and Aerospace Engineering, Oklahoma State University}
\address[adilsaddress]{Department of Engineering Cybernetics, Norwegian University of Science and Technology}

%\doublespacing
\begin{abstract}
In this paper, we introduce a decentralized digital twin (DDT) framework for dynamical systems and discuss the prospects of the DDT modeling paradigm in computational science and engineering applications. The DDT approach is built on a federated learning concept, a branch of machine learning that encourages knowledge sharing without sharing the actual data. This approach enables clients to collaboratively learn an aggregated model while keeping all the training data on each client. We demonstrate the feasibility of the DDT framework with various dynamical systems, which are often considered prototypes for modeling complex transport phenomena in spatiotemporally extended systems. Our results indicate that federated machine learning might be a key enabler for designing highly accurate decentralized digital twins in complex nonlinear spatiotemporal systems. 
\end{abstract}

\end{frontmatter}
%\linenumbers
%\doublespacing

\section{Introduction}
\label{sec:introduction}
A recent multifaceted trend in science, engineering and technology, both in academia and industry, is digital transformation \citep{westerman2014nine,arts2015digital}. Thanks to the availability of inexpensive sensors and communication technologies, improved interoperability between sensors, the phenomenal success of machine learning and artificial intelligence, in particular, deep learning, new developments in the computational hardware, and cloud and edge computing infrastructure, each and every sector is steadily moving from data-sparse regimes to the data-rich regimes. However, this growth has resulted in several major challenges. First, the data exhibits a high correlation in many systems, which implies more data does not necessarily translate into more information. Second, it is computationally demanding to train an end-to-end data-driven machine learning model that can be trustworthily used in future predictions. Third, there is a growing interest in the legal and regulatory landscape relevant to data sharing, security, privacy, and intellectual property rights (IPRs). 

One outcome of rapid digitalization is the emergence of technologies like digital twins. Broadly speaking, a digital twin can be defined as a computational model that interacts with the physical assets, and evolves over time to persistently represent the structure and behavior of the involved assets and processes \citep{kapteyn2021probabilistic,san2021hybrid,rasheed2020digital}. The digital twin concept has garnered a lot of attention with a multibillion market capitalization value for years to come \citep{san2021digital}, involving many different stakeholders all the way from the development to decommissioning phases of underlying systems and processes. Numerous disruptive concepts and start-ups have been continuously evolving to bring more value and impact to society. According to a recent report by \cite{researchdive}, the global digital twin market is predicted to grow at a compound annual growth rate of 40\%, thereby garnering \$125 billion by 2030. In particular, the availability of highly modular open-source libraries such as TensorFlow, PyTorch, and Theano, and the flexibility of cross-platforms such as Unity and Unreal Engine are lowering the barriers for many different potential use cases from precision agriculture \citep{gebbers2010precision} to precision medicine \citep{ashley2016towards} and beyond. This open-source ecosystem has allowed scientists and engineers to expose themselves faster to the state of the art digital twin technologies with cross-functionalities.       

One of the key enablers of digital twins is modeling. In computational science and engineering applications, most modeling approaches lie in either of the two categories: first principle physics-based modeling (PBM) or data-driven modeling (DDM). More recently Hybrid Analysis and Modeling (HAM) \citep{san2021hybrid} which is a fusion between the PBM and DDM is fast evolving (e.g., see also \citet{karpatne2017theory,childs2019embedding,reichstein2019deep,de2019deep,willard2020integrating,rai2020driven,sansana2021recent,sharma2022hybrid,hoffer2022theory,rajulapati2022integration}). Out of the three modeling approaches, DDM and HAM thrive on the availability of large amount of high quality data. The data requirement in the context of digital twins poses several challenges specially as one moves towards creating digital twins of increasingly complex systems. We enumerate the challenges which are most relevant in the context of the current work:

\begin{itemize}
    \item Any reasonably complex asset of a physical system or a process consists of multiple sub-components operated and maintained by different players who might have competing interests making data sharing difficult. This creates the data silo issues. 
    \item Even when the IPRs involving data is sufficiently resolved, the shared big data characterized by 5Vs (large volume, velocity, veracity, variety, value) from different components/vendors can become overwhelmingly large that might require an exorbitant amount of compute power in one place to generate knowledge.  
    \item Furthermore, the analysis of a large variety of big data in a centralized approach requires a wide array of expertise in one place which will be difficult to achieve resulting in sub-par analysis. 
    \item Lastly, the limited bandwidth of data transmission in a centralized data analysis approach limits the amount of shared data. 
\end{itemize}

To that end, the concept of federated learning can be utilized. This concept is a branch of machine learning that deals with training a model across multiple decentralized edge devices (or servers) without exchanging local data samples between them \citep{yang2019federated,zhang2021survey,wahab2021federated,liu2022distributed}. Although such privacy preserving decentralized and collaborative machine learning applications become more of a main stream approach in advertising, financial and many other industries with personalized dominance, it is relatively uncharted in advancing functionalities and providing technical support to forge a cross-domain, cross-data and cross-enterprise digital twin ecosystem. Federated machine learning algorithms have been successfully used for training machine learning models (e.g., see \citet{zhang2021survey}). In this study, we aim to extend the prospects of this concept to complex dynamical systems in order to provide building blocks (from a modeling perspective) for next generation decentralized solvers. With this in mind, our chief motivation is to communicate to a wide variety of audience on how federated machine learning can address many of the issues raised above. Consequently, this is more of a concept paper and we omit providing rigorous derivations of the utilized algorithms. Consistently, several relatively simple (but representative) problems have been chosen to demonstrate the framework. The plain vanilla versions of learning models have been utilized without any particular efforts dedicated to rigorous hyperparameter tuning and model calibration. Rather our focus is laser sharp on highlighting the exploitation of the benefits of federated machine learning in the context of digital twin.

This paper is organized as follows. Section~\ref{sec:theory} provides a brief overview on the federated learning in the context of digital twins. In Section~\ref{sec:demonstraitoncase} we detail the federated learning process for our demonstration cases. Then we validate our frameworks in Section~\ref{sec:resultsanddiscussion} demonstrating results from representative spatiotemproal dynamical systems. Our test cases include nonlinear advection-diffusion, chaotic travelling wave, and sea surface temperature forecasting processes. Finally, we highlighted our concluding remarks and outline the future research direction in Section~\ref{sec:conc}.  

\section{Federated learning in the context of digital twins}
\label{sec:theory}

    \begin{figure*}[ht]
    \centering
    \includegraphics[width=1.0\linewidth]{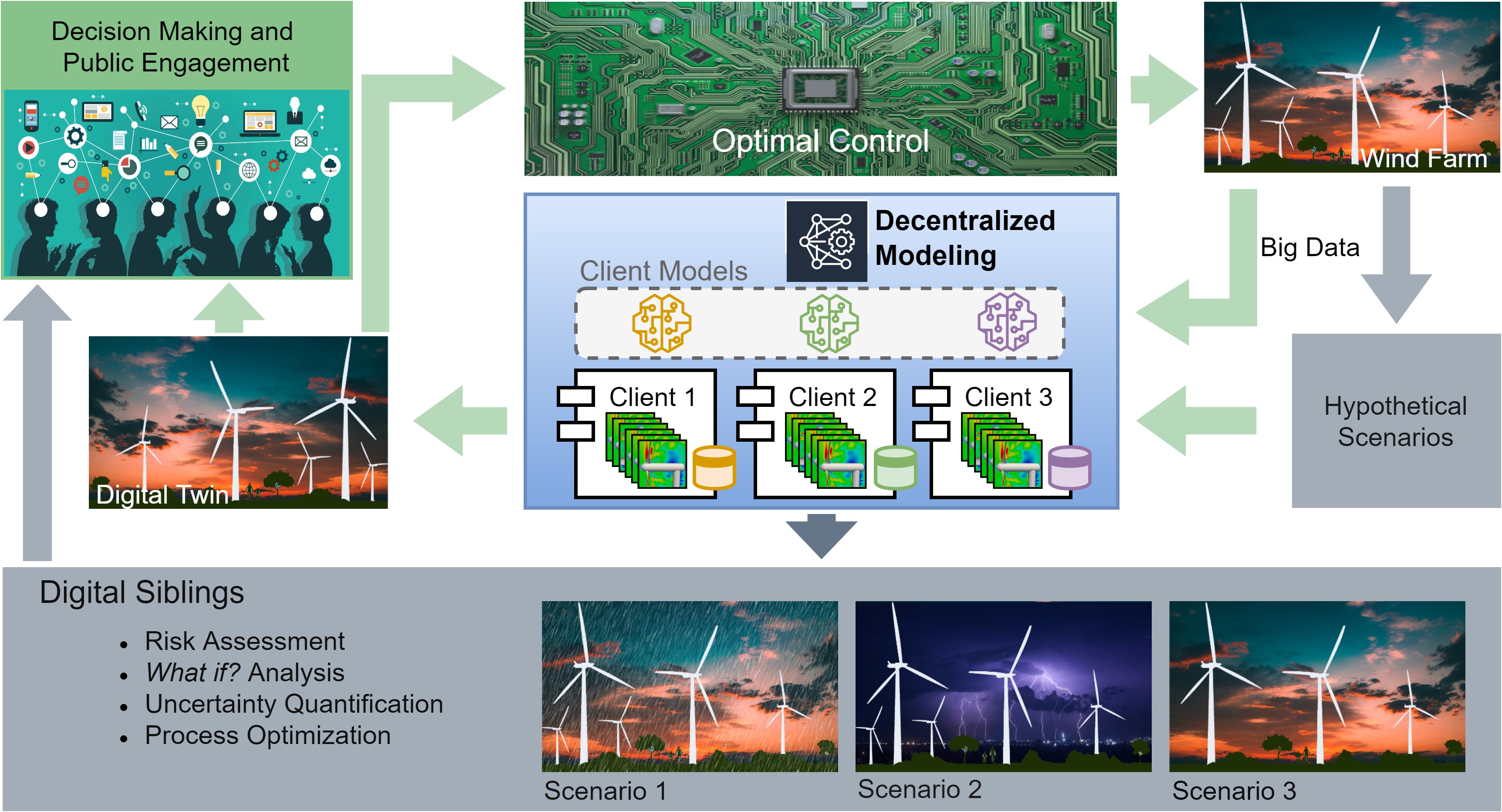}
    \caption{An overview of a decentralized digital twin concept for wind farm applications.}
    \label{fig:dtfml}
    \end{figure*}

%INCLUDE A FIGURE COMBINING OUR TRADITIONAL DT IMAGE AND THE FEDERATED LEARNING IMAGE (now I am preparing this figure, that is excellent point).
    \subsection{Digital twins}
    \label{subsec:digitaltwin}
    The need for digitalization has achieved even higher importance since the outbreak of COVID-19. The ability to control physical assets remotely under the ``work from home" situation was never felt so deeply. As the world has gradually recovered from this unfortunate situation, one can expect to see a rapid expansion of \emph{Digital Twins} \citep{rasheed2020digital}, which can be defined as virtual representations of physical assets that enable, by combining data and simulators, to perform real-time prediction, optimization, monitoring, control, and improved decision making for the physical assets. Recent advances in computational pipelines, multiphysics solvers, artificial intelligence, cybernetics, data processing and management tools are bringing now the promise of digital twins and their impact on society closer to reality. For these reasons, \emph{digital twinning} is now an important and emerging trend in many sectors, and there can be no doubt it is bound to play an even more transformative role in the post-COVID-19 world than foreseen previously. The concept of digital twins is quite rich and may benefit from many decades of research and developments in computational science and engineering.

    Digital twin, defined as a virtual representation of a physical asset enabled through data and simulators for real-time prediction, optimization, monitoring, control, and improved decision making, is a technology penetrating every domain. Figure~\ref{fig:dtfml} describes the concept of a digital twin utilizing a decentralized modeling for physical realism. On the top right side of the figure, we have the physical asset we want to represent by a digital twin. The physical asset is often equipped with a diverse class of sensors that gives big data in real-time. This data has a very coarse spatio-temporal resolution and does not describe the future state of the asset. Therefore, to complement the measurement data, models are utilized to bring physical realism into the digital representation of the asset. Provided that the same information can be obtained from the digital twin as from the physical asset, we can utilize the digital twin for informed decision-making and optimal control of the asset. However, one might be interested in risk assessment, what-if? analysis, uncertainty quantification, and process optimization. These can be realized by running the digital twin in an offline setting for scenario analysis. The concept is then known as a digital sibling. Additionally, the digital twin predictions can be archived during the asset's lifetime and can then be used for designing the next generation of assets, in which case the concept is referred to as digital threads. Digital twins, based on their capability levels, can be categorized on a scale from 0 to 5 as standalone, descriptive, diagnostic, predictive, prescriptive, and autonomous. It is evident that modeling plays a vital role in enhancing the capability of digital twins. As pointed out in \cite{san2021hybrid}, any modeling approach should at least be generalizable, trustworthy, computationally efficient, accurate, and self-adapting. Most of the modeling techniques currently lie in either the physics-based modeling or data-driven modeling categories.

    Since a digital twin requires a real time exchange of data between the physical asset and its digital representation, data privacy and security objectives become an increasingly important topic. Due to involvement of multiple stakeholders all the way from the development, through operation to the decommissioning phase of physical assets, the gathered data has to be secure both in terms of validity and privacy. Many digital twin systems and applications might involve different original equipment manufacturers, and they might want to keep their proprietary rights to secure their designs. %and expand their market to analysis and operating software.  

    The predictive digital twins are now coined with precision sciences, such as precision agriculture \citep{gebbers2010precision}, precision medicine \citep{ashley2016towards} as well as precision meteorology \citep{chilson2019moving}, in emphasizing their improved and localized forecasting capabilities. For example, we emphasize the penetration of data-driven modeling in precision meteorology, which often refers to accurate microscale weather forecasting that is of great importance in enabling solutions to water, food, energy and climate challenges in the coming century \citep{tran2020gans}. Therefore, an increase of parameterization capabilities and characteristics along with enhanced in situ observations within the atmospheric boundary layer becomes a key concept to improve the accuracy of such microscale weather forecasting models \citep{nolan2009evaluation}. %Moreover, the very basic approach of modern numerical weather prediction is to initialize these parameterized models with observed data using data assimilation methods \citep{pawar2021data} at a given time and use the physics-based primitive equations to estimate the state of the system in the future. 

    Our chief motivation in this study relies on the fact that a greater than ever penetration of smart devices (e.g., smart weather stations, smartphones, and smartwatches) has remained an uncharted technology in many spatiotemporal extended systems, and crowdsourcing data-driven modeling could be a key enabler toward their emerging digital twin applications. For example, by 2025 there will be more than 7 billion smartphones in the world. This number is significantly huge compared to the paltry (over 10,000) official meteorological stations located around the world \citep{mildrexler2011global}.  While the analysis and utilization of data only from a few edge devices might not yield accurate predictions, the processing of data from many smart and connected devices, equipped with sensors, might be a game-changer in the weather monitoring and prediction.  In their recent report, \citet{o2021service} highlighted that the Weather Company utilizes data from over 250,000 personal weather stations. Moreover, \citet{chapman2017can} discussed how the crowdsourcing data-driven modeling paradigm can take meteorological science to a new level using smart Netatmo weather stations. As more attention shifts to smart and connected \emph{internet of things} devices, security and privacy implications of such smart weather stations have been also discussed \citep{sivaraman2018smart}. 

    \subsection{Federated learning}
    \label{subsec:federatedlearning}
    \label{sec:fml}

    \begin{figure*}[!htb]
    \centering
    \includegraphics[width=1.0\linewidth]{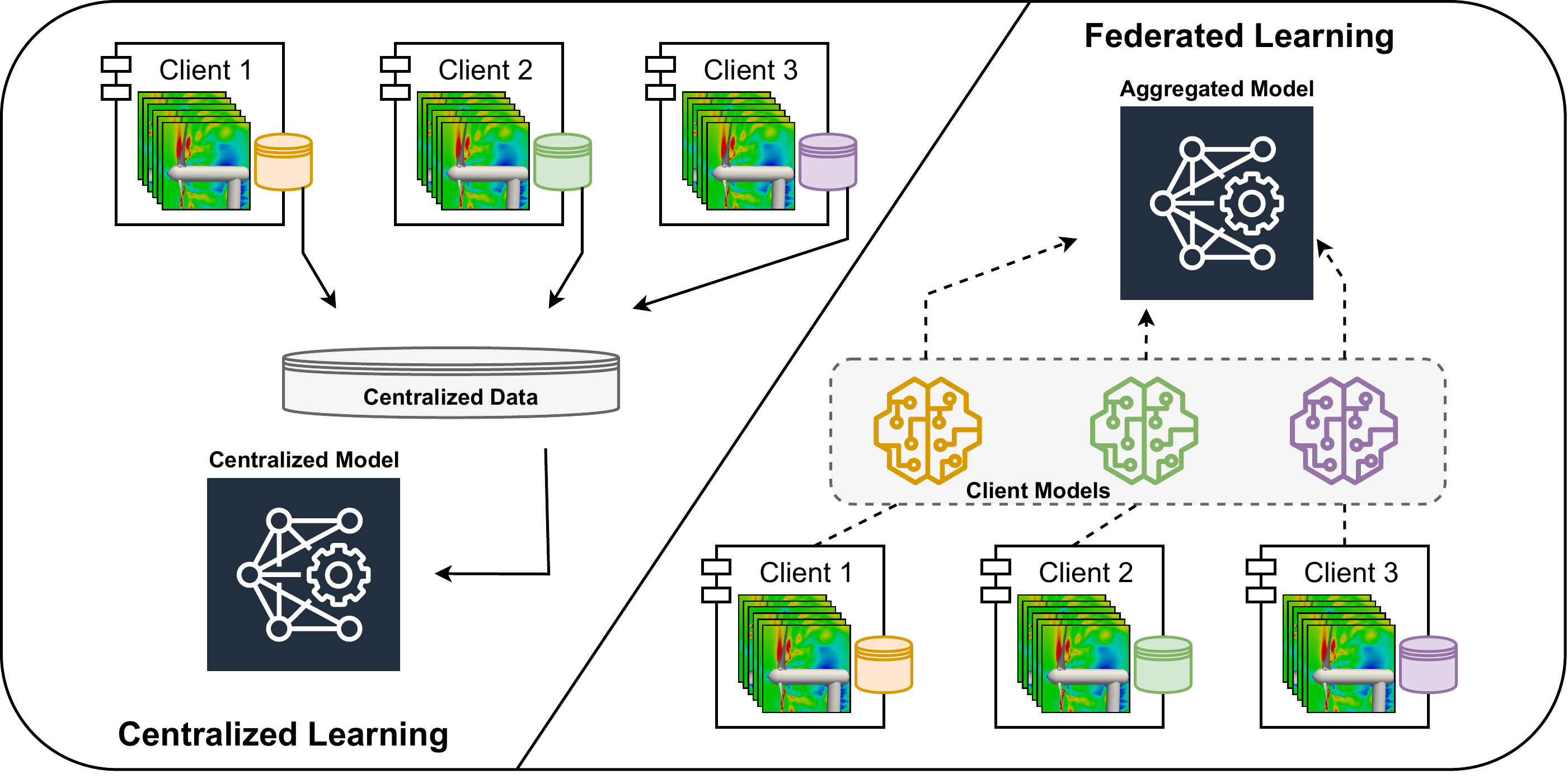}
    \caption{Overview and schematic illustrations of the centralized and federated machine learning approaches.}
    \label{fig:fml}
    \end{figure*}
    
Federated machine learning refers to the collaborative training among multiple clients without sharing the actual data. We closely follow the seminal work in federated learning \citep{mcmahan2017communication}, which introduces a federated averaging algorithm where clients collaboratively train a shared model. Figure~\ref{fig:fml} contrasts the federated learning approach with the centralized method. In the centralized method, the local dataset is transferred from clients to a central server and the model is trained using centrally stored data. In case of the federated learning, the local dataset is never transferred from clients to a server. Instead, each client computes an update to the global model maintained by the server based on the local dataset, and only this update to the model is communicated. The federated averaging algorithm assumes that there is a fixed set of $K$ clients with a fixed local dataset and a synchronous update scheme is applied in rounds of communications. At the beginning of each communication round, the central server sends the global state of the model (i.e., the current model parameters) to each of the clients. Each client computes the update to the global model based on the global state and local dataset and this update is sent to a server. The server then updates the global state of the model based on the local updates received from all clients, and this process continues. The objective function for a federated averaging algorithm can be written as follows 
    \begin{equation}
        f(w) = \sum_{k=1}^{K} \frac{n_k}{n}F_k(w) \quad \text{where} \quad F_k(w)=\frac{1}{n_k} \sum_{i \in \mathcal{P}_k} f_i(w),
    \end{equation}
    where $\mathcal{P}_k$ is the data on the $k$th client,  $n_k$ is the cardinality of $\mathcal{P}_k$, $f_i(w)=l(x_i,y_i;w)$ is the loss of the prediction on example $(x_i,y_i)$, and $w$ refers to the trainable model's parameters. The above aggregation protocol can be applied to any machine learning algorithm. 
    %In this work, we use the autoencoder for nonlinear dimensionality reduction, and 
    A complete pseudo-code for deep learning models in a federated learning setting is provided in Algorithm~\ref{alg:fl}. 
    We highlight that the approach we utilize in our study simply weights edge devices proportionally by the data they own. More advanced approaches can be considered to mitigate such limitations \citep{li2020federated,li2019fair,fallah2020personalized,deng2020adaptive,tan2022towards}, but that is beyond the scope of this paper.

    \begin{algorithm}
    \caption{Federated averaging algorithm. $B$ is the local minibatch size, $E$ is the number of local epochs, and $\alpha$ is the learning rate, and $w$ refers to the trainable parameters. }\label{alg:fl}
    \begin{algorithmic}[]
    \server{}
    \State initialize $w_0$
    \For{$t=1,2,\dots$}
    \For{each client k}
    \State{$w_{t+1}^k \leftarrow$ ClientUpdate($k,w_t$)} \Comment{update for client $k$}
    \EndFor
    \State{$w_{t+1} \leftarrow \sum_{k=1}^K\frac{n_k}{n} w_{t+1}^k$} \Comment{weight clients proportionality by the data they own}
    \EndFor
    \vspace{1em}
    \client{}
    \State{$\mathcal{B} \leftarrow $ (split $\mathcal{P}_k$ into batches of size B)}
    \For{each local epoch $i$ from 1 to $E$}
    \For{batch $b \in \mathcal{B}$}
    \State{$w \leftarrow w - \alpha \nabla l(w;b)$}
    \EndFor
    \EndFor
    \State{return $w$ to a server}
    \end{algorithmic}
    \end{algorithm}
    
\section{Demonstration cases}
\label{sec:demonstraitoncase}
To demonstrate the value of the federated learning we chose the following three cases. 
    \subsection{Burgers system}
\begin{figure}[!htb]
    \centering
    \includegraphics[width=1.0\linewidth]{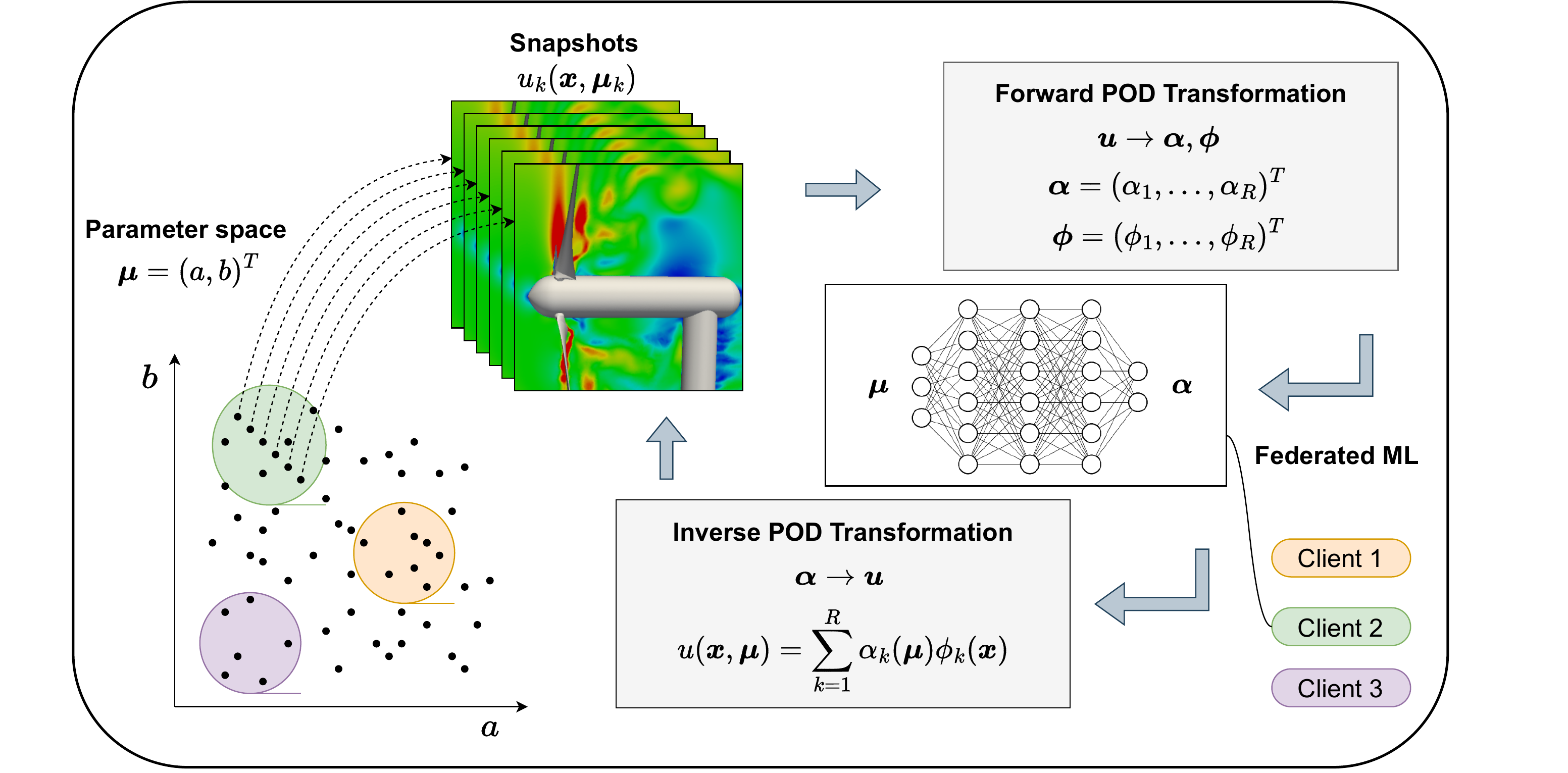}
    \caption{Federated reduced order modeling framework. This approach enables clients to collaboratively learn an aggregated model while keeping all the training data on each client. In other words, it encourages knowledge sharing without sharing the actual data. It deals with training a model across multiple decentralized edge devices (or servers) without exchanging local data samples between them.}
    \label{fig:rom}
    \end{figure}

    Our first test case for demonstration of the DDT framework is the one-dimensional Burgers equation, which is a prototypical example for nonlinear advection-diffusion problems. The Burgers equation can be written as follows
    \begin{equation}
        \dfrac{\partial u}{\partial t} + u \dfrac{\partial u}{\partial x} = {\nu} \dfrac{\partial ^2 u}{\partial x^2}, \quad x \in [0,1], 
        \label{eq:burgers}
    \end{equation}
    where $\boldsymbol{u}(\boldsymbol{x}) \in \mathbb{R}^{N_x}$ is the velocity field. We can write the analytical solution of the Burgers equation as follows
    \begin{equation}
        u(x,t) = \frac{\frac{x}{t+1}}{1 + \sqrt{\frac{t+1}{t_0}} \text{exp} \big( \frac{x^2}{4\nu(t + 1)} \big)},
    \end{equation}
    where $t_0=\text{exp}(1/8\nu)$. We parameterize the Burgers equation using the time $t$ and viscosity $\nu$ as the parameter space, i.e., $\boldsymbol{\mu}=(t,\nu)^T$. The first step of the reduced order modeling (ROM) is to generate the data. We generate the data for different values of parameters. Specifically, the parameter $t$ is varied between 0 to 2, and the parameter $\nu$ lies between 0.001 to 0.01. Once the data is generated, we form the matrix $\mathbf{A}$ $\in$ $\mathbb{R}^{N_x \times N}$ whose columns are the $\boldsymbol{u}_n$ corresponding to $\boldsymbol{\mu}_n$, and then perform the singular value decomposition (SVD) of the matrix
    \begin{equation}
        \mathbf{A} = \mathbf{W}\mathbf{\Sigma}\mathbf{V}^T = \sum_{k=1}^N\sigma_k \mathbf{w}_k \mathbf{v}_k^T,
    \end{equation}
    where $\mathbf{W}$ is an $N_x \times N$ matrix with orthonormal columns $\mathbf{w}_k$, $\mathbf{V}$ is an $N \times N$ matrix with orthonormal columns $\mathbf{v}_k$, and $\mathbf{\Sigma}$ is an $N \times N$ matrix with non-negative diagonal entries, called singular values, arranged such that $\sigma_1 \ge \sigma_2 \ge \dots \ge \sigma_N \ge 0$. The vectors $\mathbf{w}_k$ are the proper orthogonal decomposition (POD) modes that we denote as $\phi_k$, and $\mathbf{\Phi}=\{ \phi_k\}_{k=1}^{R}$ is the set of POD basis functions \citep{rowley2017model}. The representation of the approximated velocity field using the POD modes is as follows
    \begin{equation} \label{eq:ipod}
        \boldsymbol{u}(\boldsymbol{x}, \boldsymbol{\mu}) = \sum_{k=1}^{R} \alpha_k(\boldsymbol{\mu}) \phi_k(\boldsymbol{x}),
    \end{equation}
    where $\alpha_k{(\boldsymbol{\mu})}$ are the modal coefficients corresponding to parameter $\boldsymbol{\mu}$, and $R$ refers to the number of modes retained in our model. The neural network is trained to learn the relationship between the parameters of the Burgers equation to the POD modal coefficients. Once the neural network is trained, we can predict the POD modal coefficients for a new set of parameters, and the velocity field can be reconstructed using Equation~\ref{eq:ipod}. In the case of the federated ROM framework, the low-dimensional data is assumed to be distributed across clients and the training is done using the federated averaging algorithm as described in Algorithm~\ref{alg:fl}. The complete federated ROM framework is illustrated in Figure~\ref{fig:rom} where clients collaboratively learn an aggregated model while keeping all the training data on their respective client. The machine learning model is a feed-forward neural network with four hidden layers and forty neurons per hidden layer. We use the relu activation function to introduce nonlinearity and the linear activation function is used at the output layer.   

    \subsection{Kuramoto–Sivashinsky system}
    The second test case pertains to federated machine learning applied to an autoencoder which is a powerful approach for obtaining the latent space on a nonlinear manifold. The autoencoder is composed of an encoder and a decoder, where the encoder maps an input to a low-dimensional latent space and the decoder performs the inverse mapping from latent space variables to the original dimension at the output. If we denote the encoder function as $\eta (w)$ and a decoder function is defined as $\xi (w)$, then we can represent the manifold learning as follows
    \begin{align}
        \eta, \xi &= \argmax_{\eta, \xi} ~\lVert \boldsymbol{u} - (\eta \circ \xi)\boldsymbol{u} \rVert, \\
        \eta &: \boldsymbol{u} \in \mathbb{R}^N \rightarrow \boldsymbol{z} \in \mathbb{R}^R, \\
        \xi &: \boldsymbol{z} \in \mathbb{R}^R \rightarrow \boldsymbol{u} \in \mathbb{R}^N,
    \end{align}
    where $\boldsymbol{z}$ represents the low dimensional latent space and $R$ is the dimensionality of the latent space.
    
    We demonstrate the nonlinear dimensionality reduction using an autoencoder for spatiotemporal chaotic systems in which local data samples are held in edge devices. The encoder and decoder both have three hidden layers and each hidden layer has 100 neurons. We use the elu activation function for hidden layers and the linear activation function is used at the output layer of both encoder and decoder.  
    
    Specifically, we consider the Kuramoto–Sivashinsky (KS) system to demonstrate the prospects of the federated learning concept. This KS system is known for its irregular or chaotic behavior \citep{laquey1975nonlinear,kuramoto1978diffusion,sivashinsky1977nonlinear,gonzalez1998identification,pathak2018model,vlachas2018data,linot2020deep,vlachas2022multiscale}, and we often formulate the KS equation with $L$-periodic boundary conditions as follows
    \begin{equation}
        \frac{\partial u}{\partial t} + \frac{\partial^4  u}{\partial x^4} + \frac{\partial^2 u}{\partial x^2} + u \frac{\partial u}{\partial x} =0,
        \label{eq:ks}
    \end{equation}
    on a spatial domain $x \in [0, L]$, where the dynamics undergo a hierarchy of bifurcations as the spatial domain size $L$ is increased, building up the chaotic behavior. Here, we perform the underlying numerical experiments with $L=22$ to generate our spatiotemporal data set. Equation~\ref{eq:ks} is solved using the fourth-order method for stiff partial differential equations \citep{kassam2005fourth,vlachas2022multiscale} with the spatial grid size of $N=64$. The random initial condition is assigned at time $t=-250$ and the solution is evolved with a time step of $2.5 \times 10^{-3}$ up to $t=0$. Using the solution at time $t=0$ as the initial condition, the KS system is evolved till $t=2500$. The data is sampled at a time step of 0.25 and these 10,000 samples are used for training and validation. For the testing purpose, the data from $t=2500$ to $t=3750$ is utilized.

    \subsection{NOAA Optimum Interpolation Sea Surface Temperature Dataset}
    Our third test case is the decentralized autoencoder for the nonlinear dimensionality reduction of the NOAA OI SST V2 analysis dataset. This dataset is obtained using in situ  (ship and buoy) and satellite SSTs plus SSTs simulated by the sea-ice cover. Before calculating the analysis, the satellite data is adjusted for biases using the method of  \citet{reynolds1988real} and \citet{reynolds1993improved}. This dataset consists of the weekly average sea surface temperature on a $1^\circ$ latitude $\times$ $1^\circ$ longitude global grid ($180 \times 360$). Even though the SST dataset exhibits a strong periodic structure due to seasonal fluctuations, complex ocean dynamics lead to rich flow physics in this dataset. This dataset has been used in number of recent studies on geophysical emulation \citep{maulik2020recurrent}, flow reconstruction \citep{callaham2019robust}, and multi-resolution dynamic mode decomposition \citep{kutz2016multiresolution}. The autoencoder is utilized for nonlinear dimensionality reduction of the NOAA OI SST V2 dataset. The autoencoder is composed of the encoder function $\eta(w)$ which maps the high dimensional sea surface temperature field to low dimensional latent space and a decoder function $\xi(w)$ which maps the low dimensional latent space to the same sea surface temperature field. Both encoder and decoder are parameterized by the weights $w$ and these parameters are learned through training. We can represent the reconstruction of the temperature field with the autoencoder as follows
    \begin{align}
        \eta, \xi &= \argmax_{\eta, \xi} ~\lVert \boldsymbol{\theta} - (\eta \circ \xi)\boldsymbol{\theta} \rVert, \\
        \eta &: \boldsymbol{\theta} \in \mathbb{R}^N \rightarrow \boldsymbol{z} \in \mathbb{R}^R, \\
        \xi &: \boldsymbol{z} \in \mathbb{R}^R \rightarrow \boldsymbol{\theta} \in \mathbb{R}^N,
    \end{align}
    where $\boldsymbol{\theta}$ is the sea surface temperature field, $\boldsymbol{z}$ represents the low dimensional latent space and $R$ is the dimensionality of the latent space at the bottleneck layer of the autoencoder. Here, we use the data from October 1981 to July 2010 (1500 snapshots) for training the autoencoder model and the data beyond July 2010 is used for comparing the performance of the trained autoencoder model with the unseen data.

\section{Results and discussion}
\label{sec:resultsanddiscussion}
    \subsection{A decentralized reduced order modeling framework: Burgers system}
    \label{sec:burgers}

    \begin{figure}[!htb]
    \centering
    \includegraphics[width=0.8\linewidth]{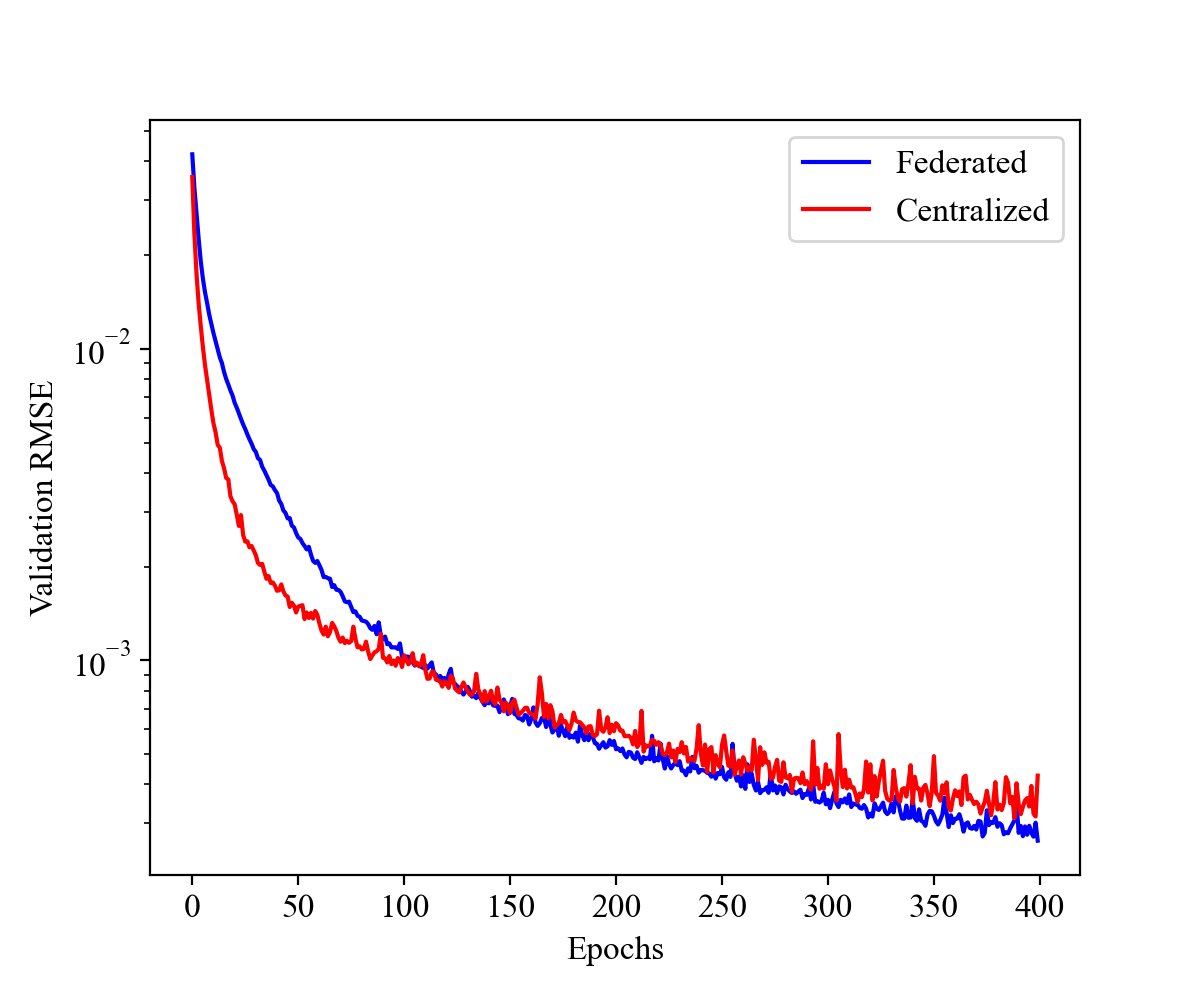}
    \caption{Validation loss during training of the centralized and federated neural network for the ROM framework of Burgers equation.}
    \label{fig:mse}
    \end{figure}
    
    \begin{figure}[!htb]
    \centering
    \includegraphics[width=1.0\linewidth]{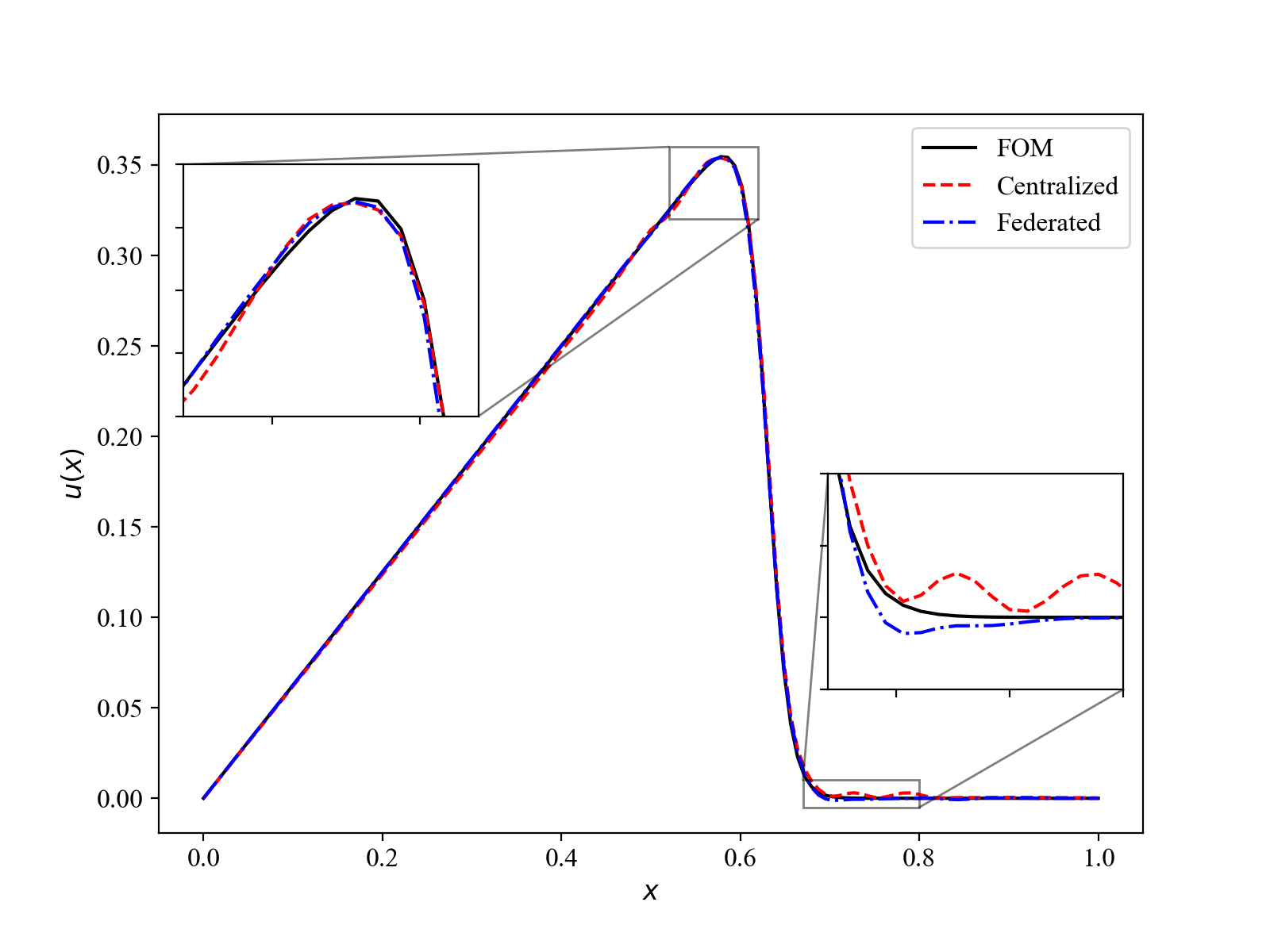}
    \caption{Reconstruction performance of the centralized and federated learning approaches for Burgers equation for $\boldsymbol{\mu}=(0.02, 0.00475)^T$.}
    \label{fig:rec_burgers}
    \end{figure}
    
    The mean squared error loss for the validation dataset during the training of centralized and federated ROM frameworks is shown in Figure~\ref{fig:mse}. The validation loss for both centralized and federated learning follows a similar trajectory during the training indicating that there is no loss in federated learning. Therefore, federated learning can provide us with a deep learning model with a similar level of accuracy as centralized learning without the need to share the data. The performance of the parametric ROM framework for $\boldsymbol{\mu}=(0.02, 0.00475)^T$ is shown in Figure~\ref{fig:rec_burgers}. Compared against the FOM data, which is the analytical (true) solution in this example, we see that both centralized and federated ROMs are highly accurate in reconstructing the velocity field for the Burgers equation.   

    \subsection{A decentralized autoencoder framework: Kuramoto–Sivashinsky system}
    \label{sec:ks}
    \begin{figure}[!htb]
    \centering
    \mbox{
    \subfigure[$R=6$]{\includegraphics[width=0.5\textwidth]{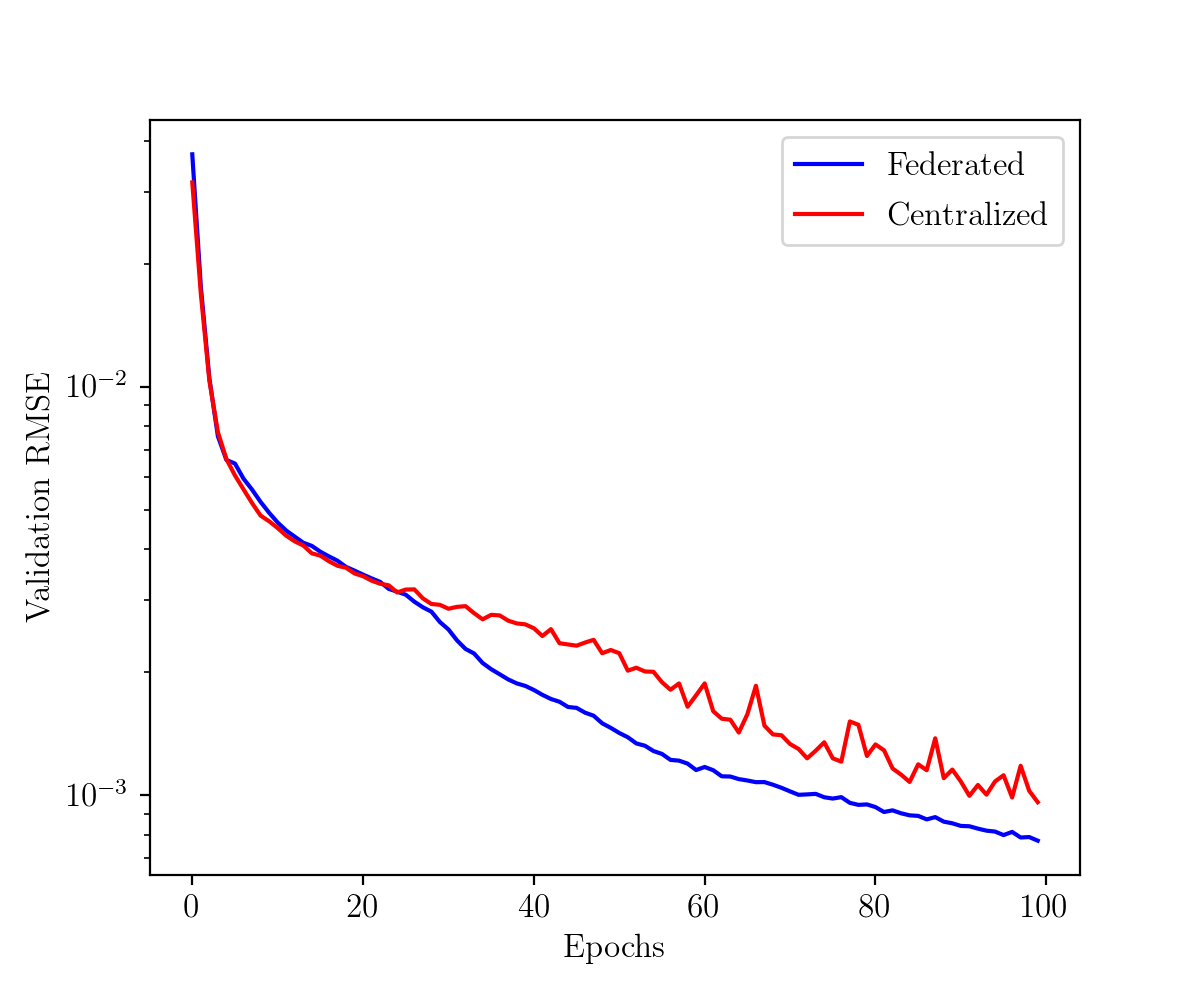}}
    \subfigure[$R=8$]{\includegraphics[width=0.5\textwidth]{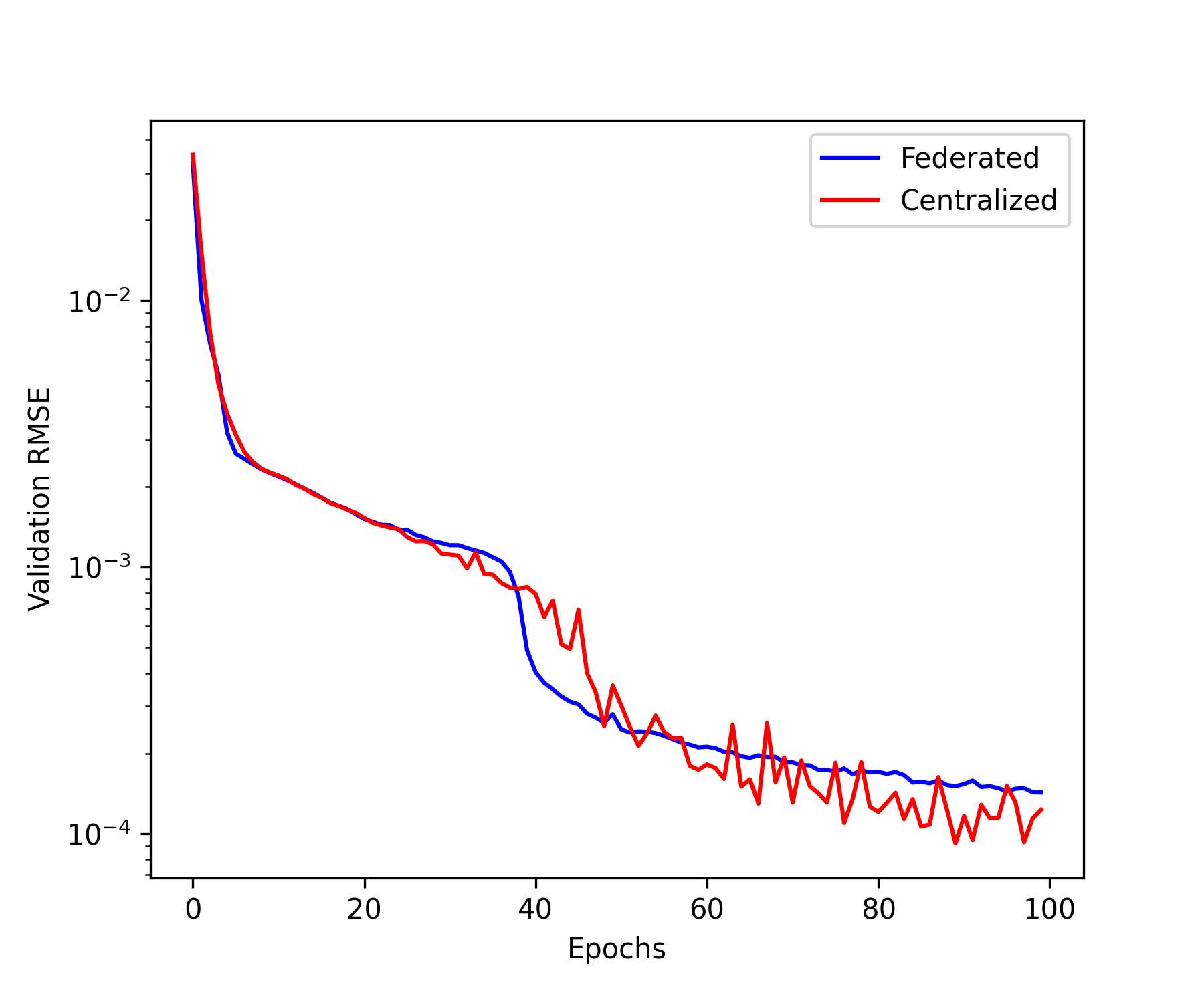}}
    }\\
    \mbox{
    \subfigure[$R=10$]{\includegraphics[width=0.5\textwidth]{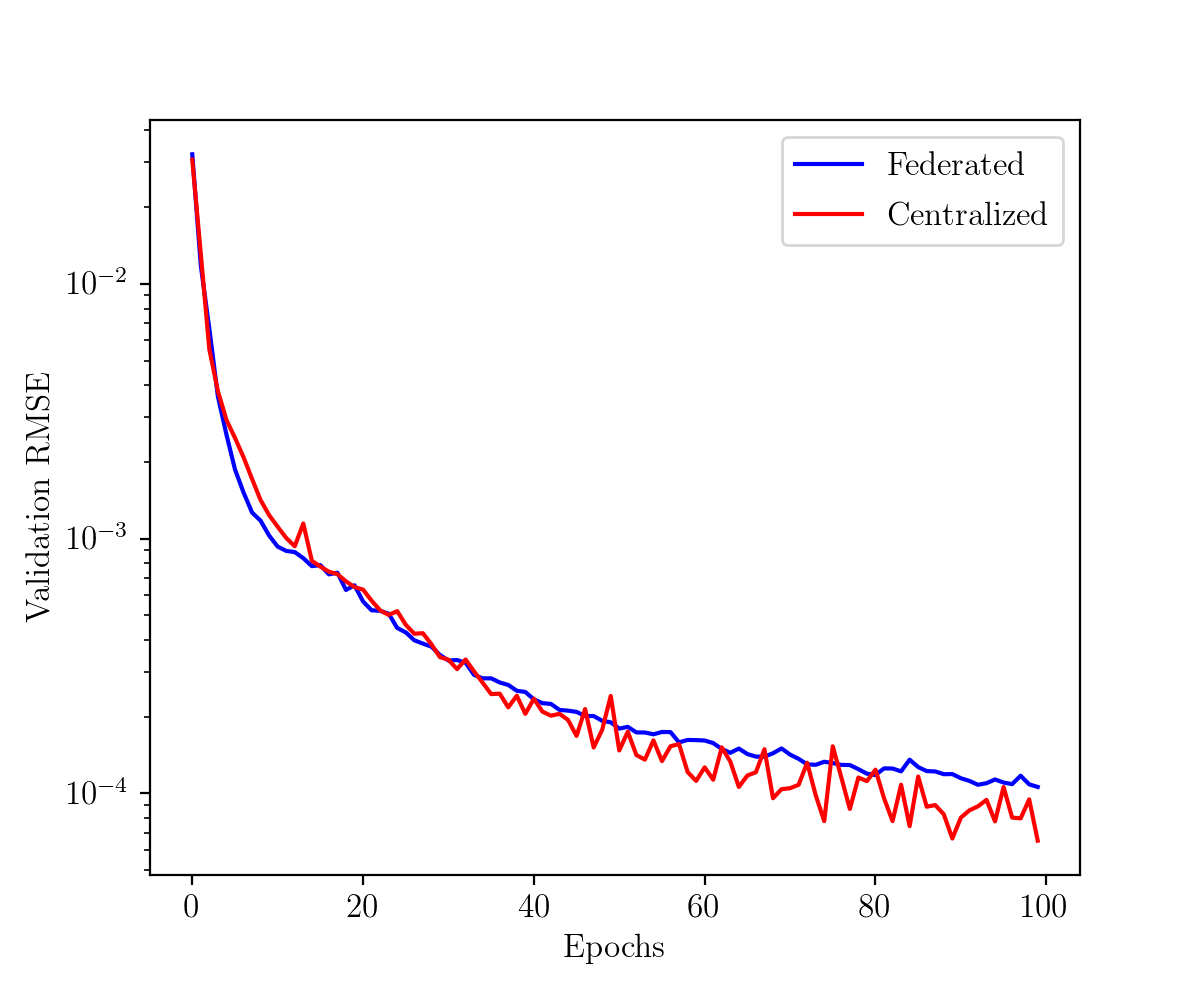}}
    \subfigure[$R=12$]{\includegraphics[width=0.5\textwidth]{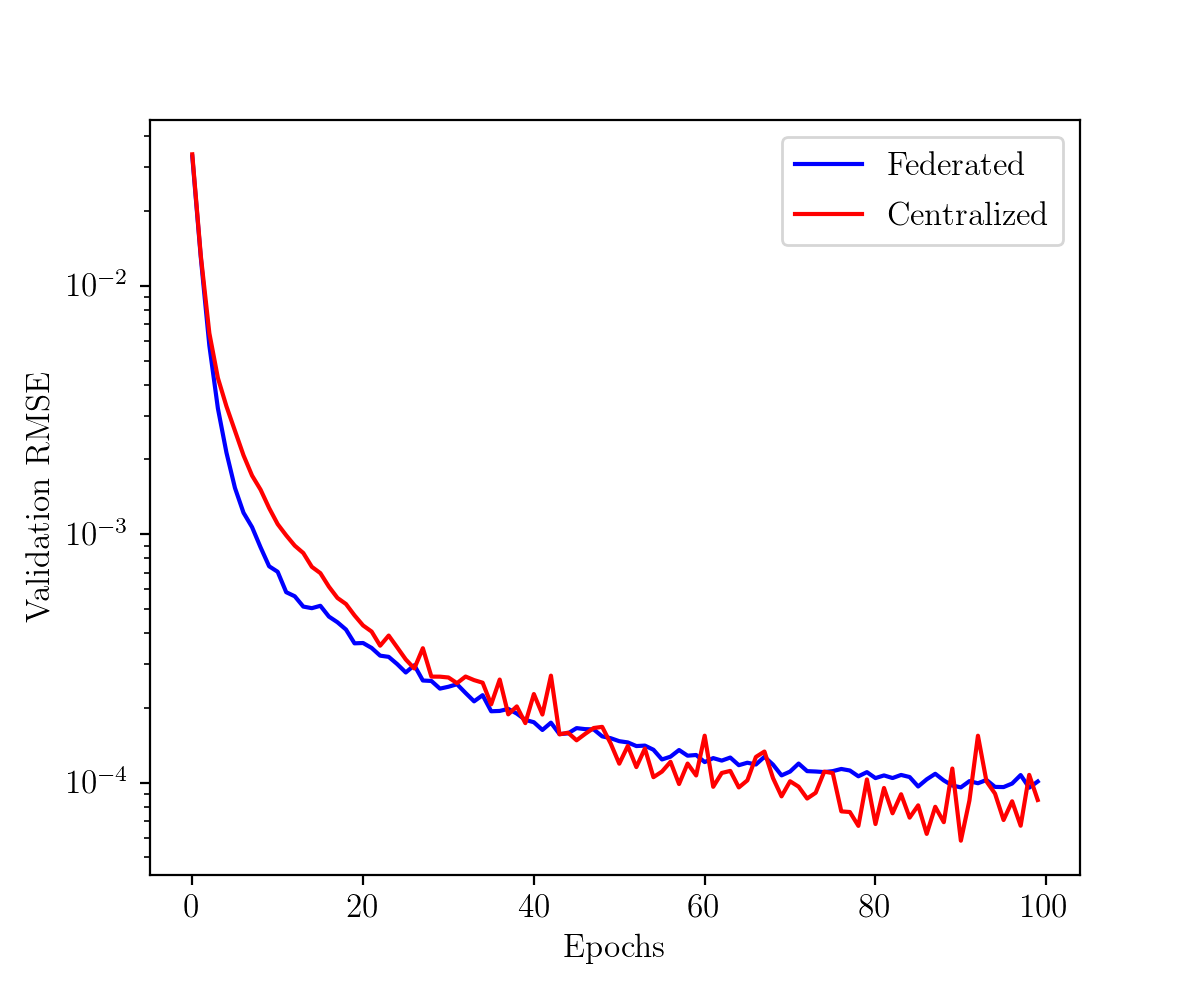}}
    }
    \caption{Validation loss during training of the centralized and federated neural network for the Kuramoto–Sivashinsky system.}
    \label{fig:ks_loss}
    \end{figure}
    
    \begin{figure}[!htb]
    \centering
    \includegraphics[width=1.0\linewidth]{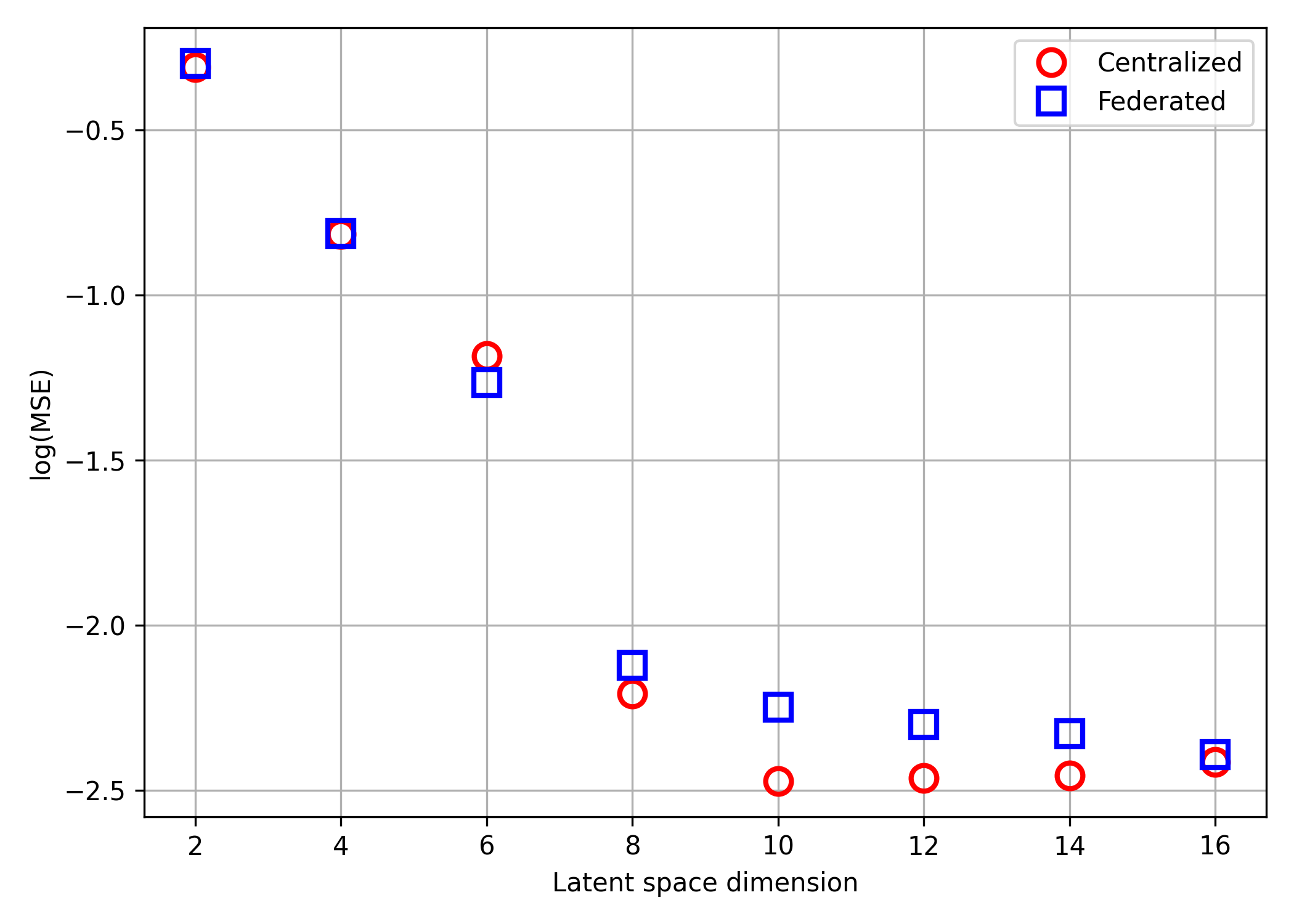}
    \caption{Mean square error representing the reconstruction performance of the federated and centralized autoencoders considering different latent space dimensions in the Kuramoto–Sivashinsky system.}
    \label{fig:ks_latent}
    \end{figure}
 
     \begin{figure}[!htb]
    \centering
    \includegraphics[width=1.0\linewidth]{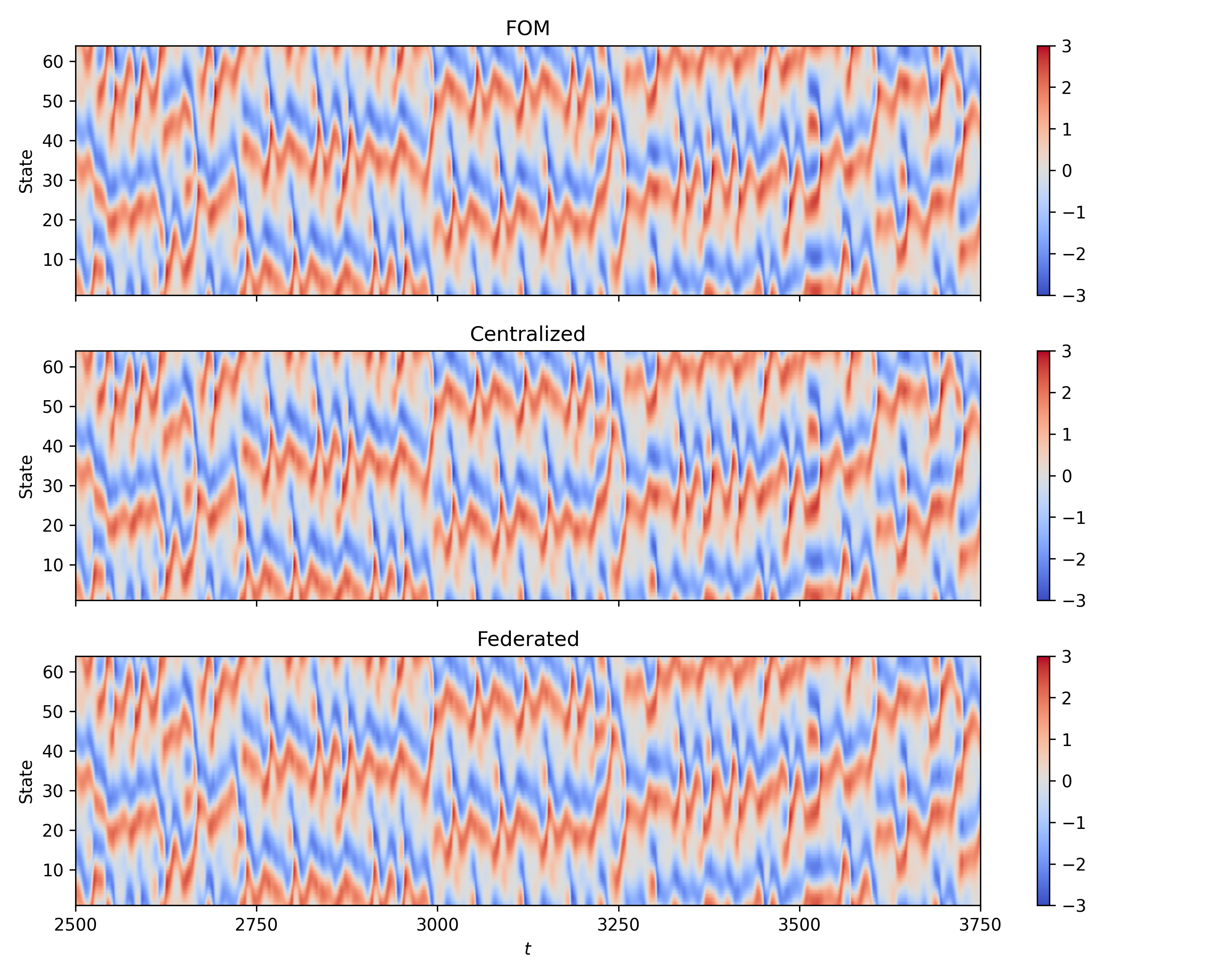}
    \caption{Space-time solutions of the Kuramoto–Sivashinsky system obtained by the federated and centralized models.}
    \label{fig:ks_rec}
    \end{figure}

    \begin{figure}[!htb]
    \centering
    \includegraphics[width=1.0\linewidth]{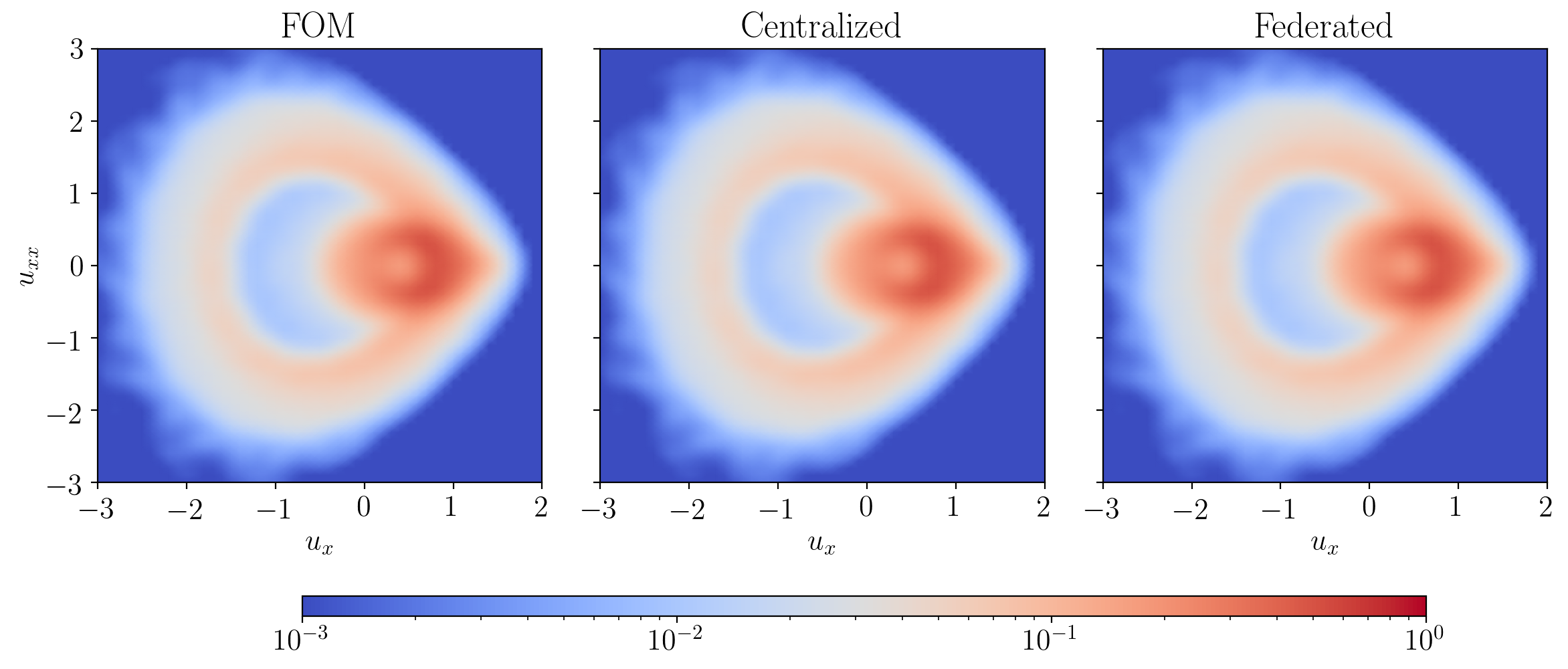}
    \caption{Joint PDF of pointwise values of $u_x$ and $u_{xx}$ on a logarithmic scale for solving the Kuramoto–Sivashinsky system.}
    \label{fig:ks_pdf}
    \end{figure}
    
    \begin{figure}[!htb]
    \centering
    \mbox{
    \subfigure[$R=10$]{\includegraphics[width=0.5\textwidth]{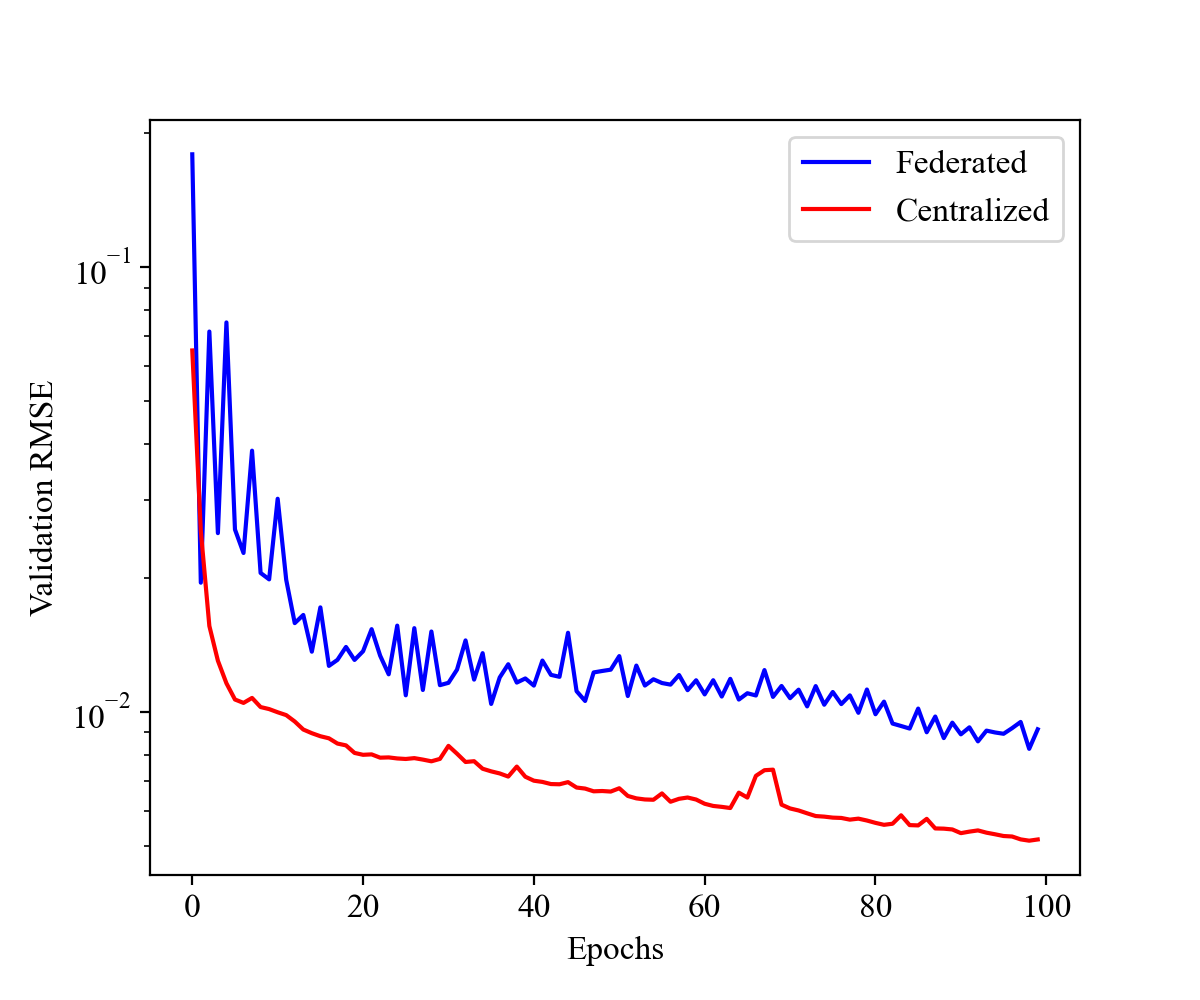}}
    \subfigure[$R=15$]{\includegraphics[width=0.5\textwidth]{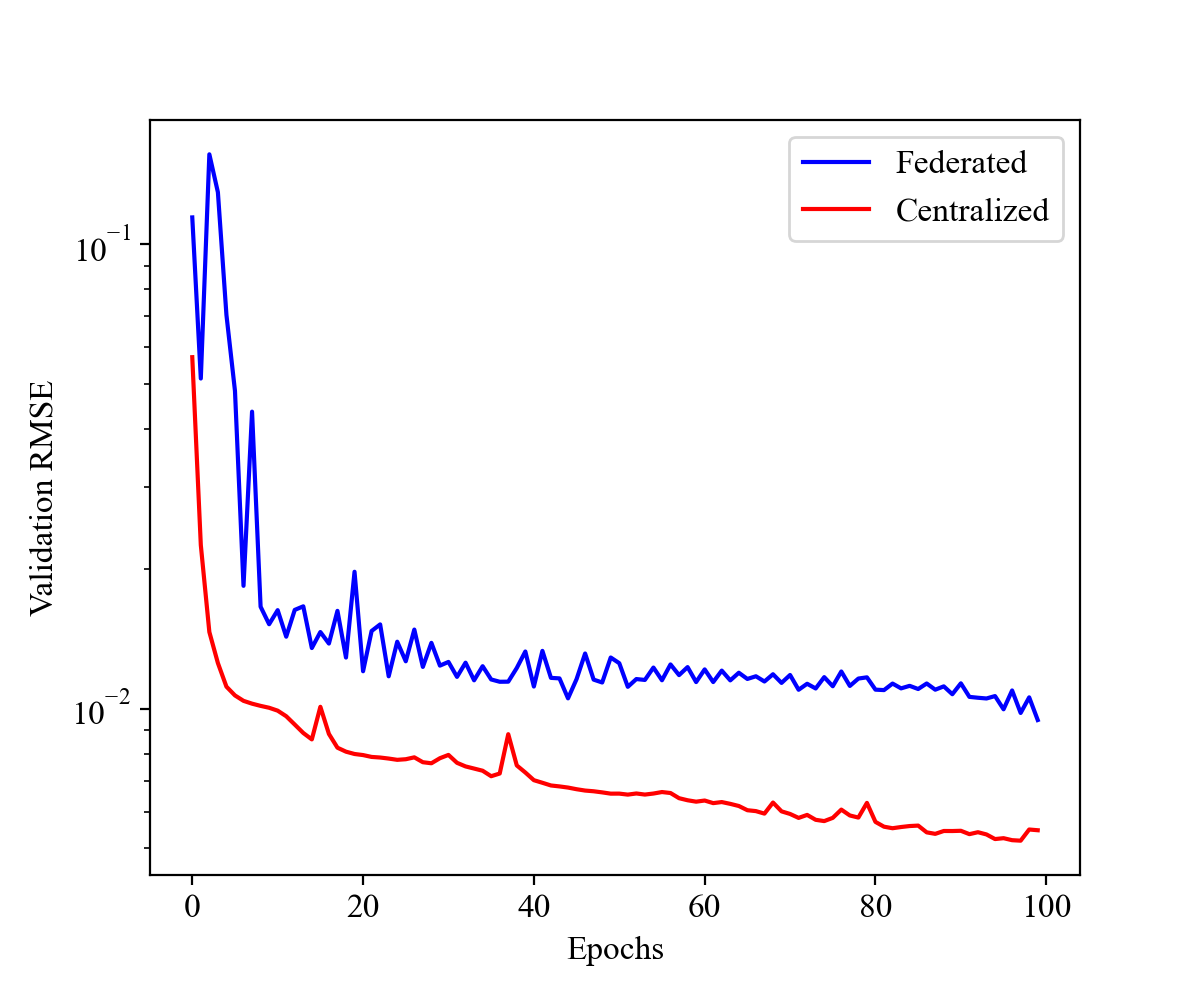}}
    }\\
    \mbox{
    \subfigure[$R=20$]{\includegraphics[width=0.5\textwidth]{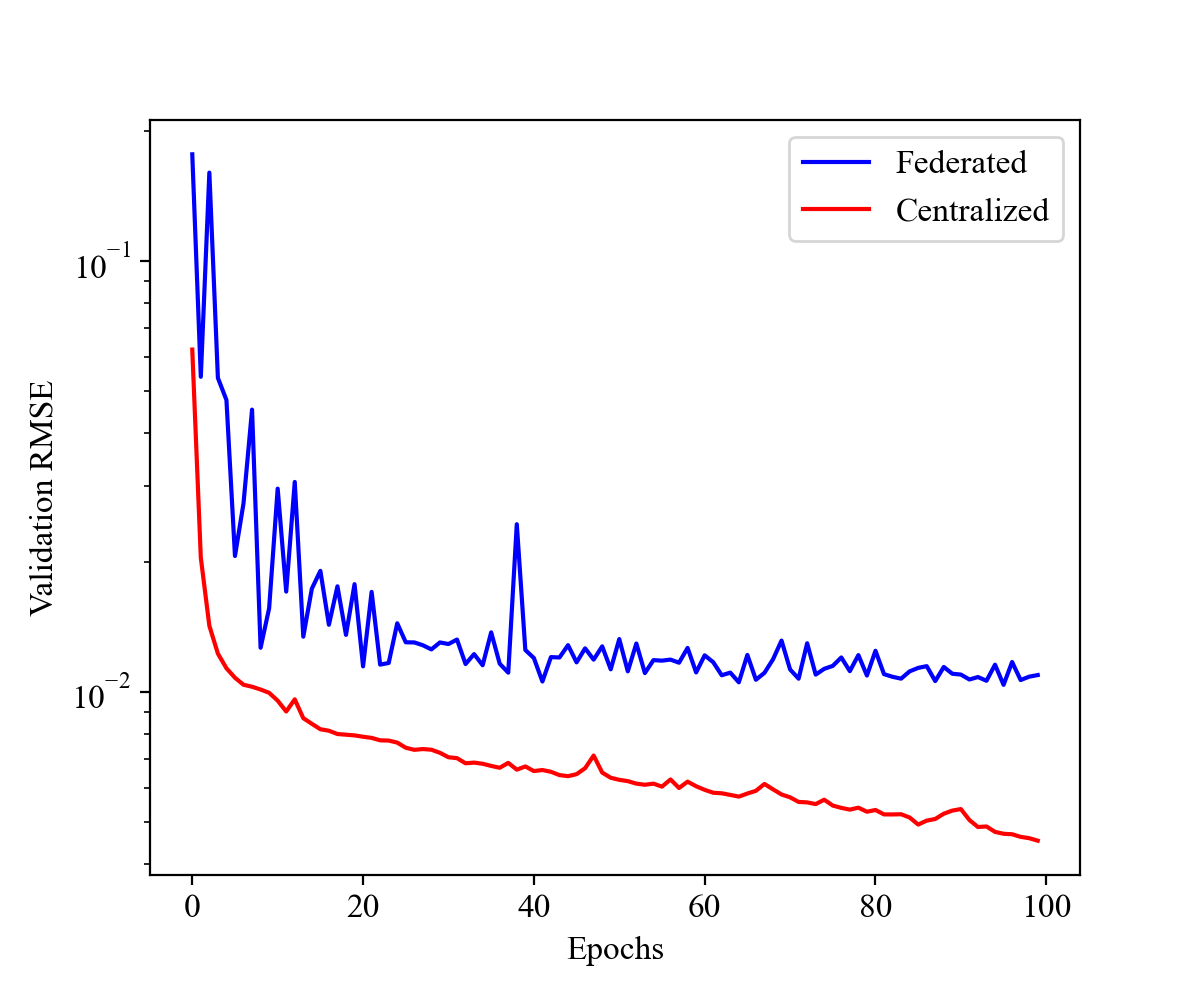}}
    \subfigure[$R=25$]{\includegraphics[width=0.5\textwidth]{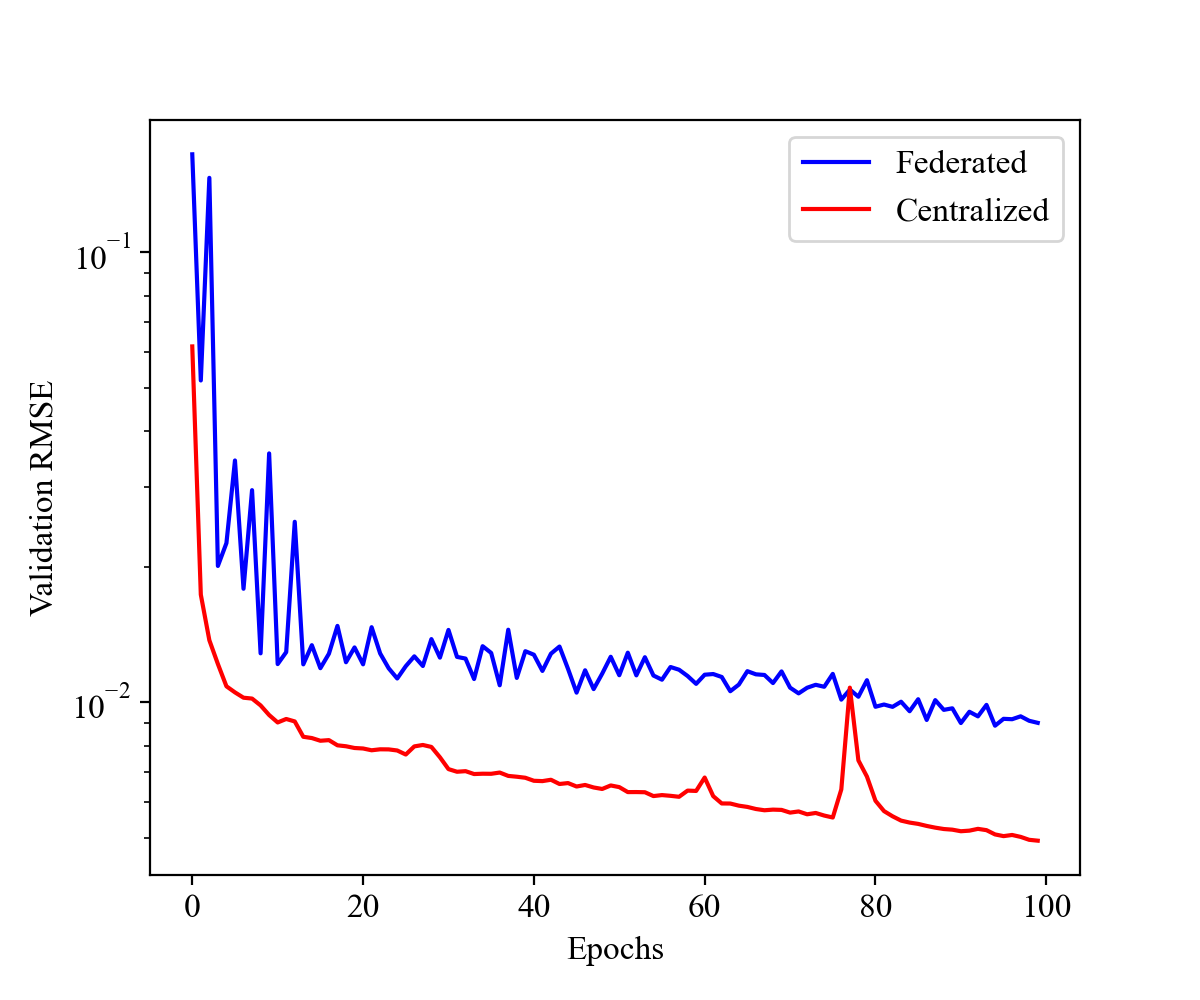}}
    }
    \caption{Validation loss during training of the centralized and federated neural network for the NOAA Optimum Interpolation Sea Surface Temperature Dataset.}
    \label{fig:noaa_loss}
    \end{figure}

    Following the work of \citet{vlachas2022multiscale}, we first validate the centralized approach by varying $R$. For the federated learning, we use $K=10$ clients, and each client model is trained for $E=1$ local epoch with a batch size $B=32$. For a fair comparison, the batch size of 320 is utilized for training the centralized autoencoder. The validation loss for the centralized and federated autoencoder with different dimensionality of the latent space is depicted in Figure~\ref{fig:ks_loss}, and we see that both the losses converge to very similar values. This shows that there is no loss in accuracy due to federated learning compared to centralized learning. As shown in Figure~\ref{fig:ks_rec}, the reconstruction error for both centralized and federated autoencoders saturates around $R=8$ modes. Our observations are consistent with previous works \citep{vlachas2022multiscale,linot2020deep,cvitanovic2010state,robinson1994inertial}, suggesting that the latent space dynamics lies effectively on a manifold with $R=8$ dimensions.

    The state trajectory of the KS system for the testing period is shown in Figure~\ref{fig:ks_rec}. The true data (i.e., FOM) is also included in the figure for comparison purpose.
    %along with the error between the true data (i.e., FOM) and reconstructed data from centralized and federated autoencoders. The error is computed as the absolute difference between the true and predicted state of the KS system. 
    Both the centralized and federated autoencoders have a similar level of predictive performance. The joint probability distribution function (PDF) of the pointwise values of $u_x$ and $u_{xx}$ is shown in Fig.~\ref{fig:ks_pdf}. We observe that the joint PDF of the prediction from both centralized and federated autoencoder matches very closely with the true joint PDF.

    \begin{figure}[!htb]
    \centering
    \includegraphics[trim={4cm 0 4cm 0},clip,width=0.9\linewidth]{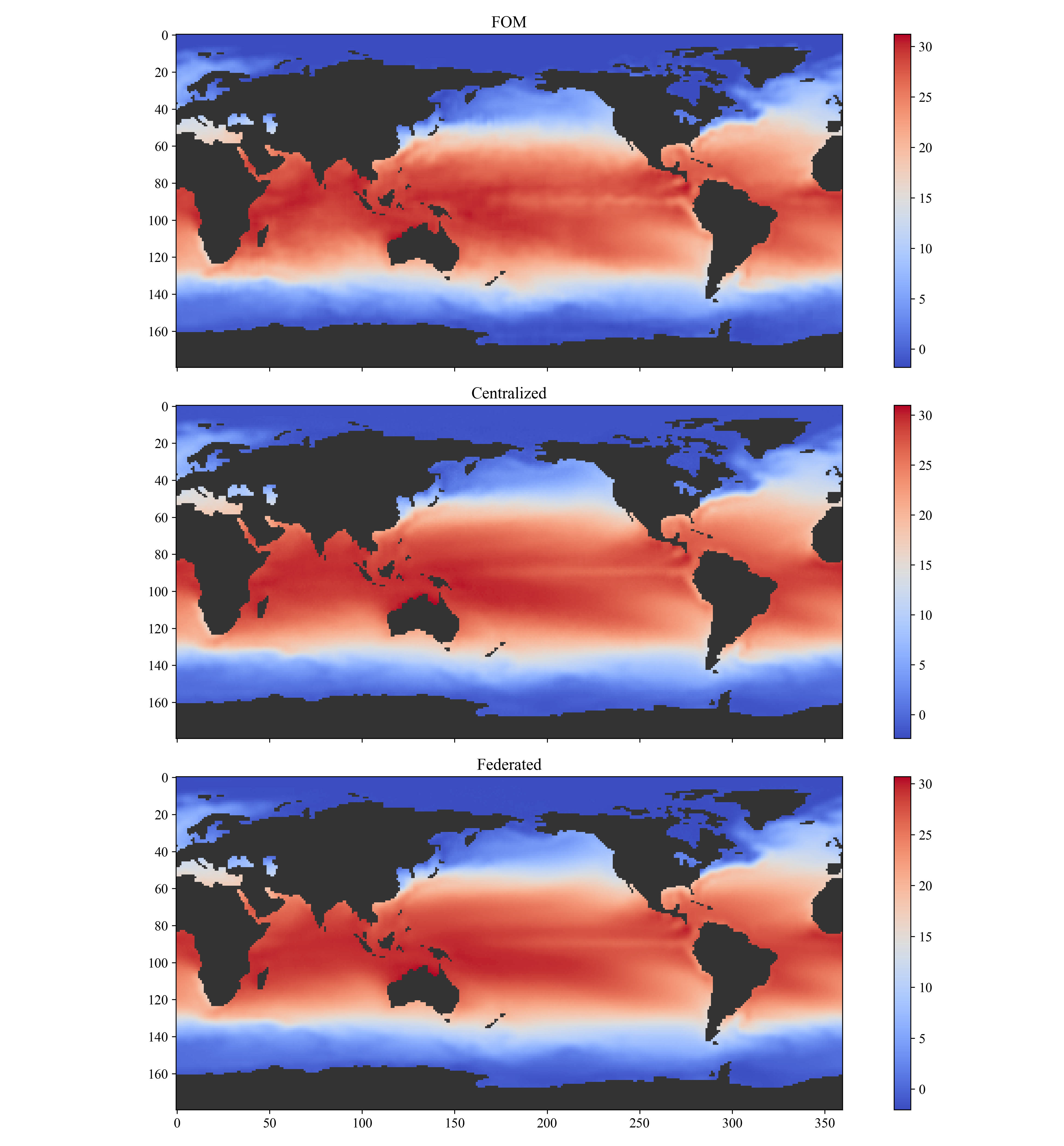}
    \caption{Weekly averaged temperature field in degrees Celsius for the fourth week of March 2018 along with reconstructed temperature field from the centralized and federated autoencoders.} 
    \label{fig:noaa_rec}
    \end{figure}

    \subsection{A decentralized autoencoder framework: NOAA Optimum Interpolation Sea Surface Temperature Dataset}
    \label{sec:sst}
    First, we show the performance of centralized and federated autoencoders for different dimensionality of the latent space at the bottleneck layer. The encoder is composed of three hidden layers with the number of neurons being $\{ 800, 400, 200\}$ across the first, the second, and the third hidden layer, respectively. The decoder has the same architecture in reverse order, i.e., three hidden layers with $\{ 200, 400, 800\}$ neurons. The number of neurons in the bottleneck layer is varied between $R=5$ to $R=25$ with an increment of $5$. The encoder and decoder architecture remain the same for all the numerical experiments. Figure~\ref{fig:noaa_loss} depicts the validation loss during both centralized and federated training of the autoencoder. The validation loss for federated learning is slightly higher compared to the centralized learning from the first epoch. This can be attributed to the fact that in federated averaging the weights of the global model are initialized by simply averaging the weights of the client model. The simple averaging of weights might not be the most effective strategy especially when the amount of the data is small for each client. Despite this, the validation loss for both centralized and federated learning follows a similar trajectory with the loss reduced by almost one order of magnitude from the initial loss. The reconstruction of the weekly average temperature field for the fourth week of March 2018 is shown in Figure~\ref{fig:noaa_rec} for both centralized and federated autoencoder along with the full order model, i.e., the analysis.

\section{Conclusion}
\label{sec:conclusion}
\label{sec:conc}
In this work, we explore the potential of federated machine learning in the context of digital twin operated by multiple vendors for modeling complex spatiotemporal dynamical systems. In particular, we investigated federated learning for nonlinear dimensionality reduction of dynamical systems. Federated learning allows for collaborative training of a model while keeping the training data decentralized. Our numerical experiments show that a federated model can achieve the same level of accuracy as the model trained using the central data collected from all clients. This work opens up the possibility of updating a global model without exposing the local data collected from different sources. In summary, we addressed the four main challenges related to data mentioned in the introduction section as follows:
\begin{itemize}
    \item Since the data generated by individual vendors never leaves the local servers, a decentralized setting guarantees better security, addresses key concerns related to IPRs, and data silo issues. Furthermore, since each vendor can focus on individual sub-component of the asset, the data quality can be significantly enhanced.
    \item Since each data owner can conduct the training using the limited amount of data that they generate and own, they can work with significantly low compute power in a cost effective way. 
    \item The decentralized federal learning approach means that specialists can be involved more effectively to analyze, interpret, and train models on the data that they have good insights into. Also, depending upon the quality associated with measuring equipment/device (The Meteorological Office approved specialized and expensive instruments vs non-professional Netatmo stations) of different stakeholders, different weights can be applied 
    \item Since the data is converted into knowledge which are generally a highly compressed representation, the knowledge transfer can taken place over very limited bandwidth. 
\end{itemize}

Although in this study we primarily focus on federated learning in the context of spatiotemporal reconstruction of dynamical systems, our approach can be also generalized to large-scale computational settings beyond transport phenomena, for which the research outcomes might improve broader modeling and simulation software capabilities to design fast, cohesive, effective, and secure predictive tools for cross-domain simulations in the various levels of information density. 

In summary, the trained models contributed by the different stakeholder can be reversed engineered thereby compromising data security. In our future studies, we plan to leverage decentralized learning approaches in the context of precision meteorology and develop new physics-guided federated learning approaches to forge new surrogate models compatible among heterogeneous computing environments. Also, we would like to investigate which federated learning approach is best to fuse the weights. The three demonstration cases handled here are relatively simple but that was completely intentional as it eases the communication and dissemination of the work to a larger audience. However, the next logical step should be to apply the approach to a more complex problem.

\section*{Acknowledgments}
%The second and third author are grateful for the support received by the Research Council of Norway and the industrial partners of the following projects: EXAIGON--{\em Explainable AI systems for gradual industry adoption\/} (grant no. 304843), {\em Hole cleaning monitoring in drilling with distributed sensors and hybrid methods\/} (grant no. 308823), and RaPiD--{\em Reciprocal Physics and Data-driven models\/} (grant no. 313909). The fourth author gratefully acknowledges the Early Career Research Program (ECRP) support of the U.S. Department of Energy, Office of Science, Office of Advanced Scientific Computing Research under Award Number DE-SC0019290.

This material is based upon work supported by the U.S. Department of Energy, Office of Science, Office of Advanced Scientific Computing Research under Award Number DE-SC0019290. O.S. gratefully acknowledges their Early Career Research Program support. 

Disclaimer: This report was prepared as an account of work sponsored by an agency of the United States Government. Neither the United States Government nor any agency thereof, nor any of their employees, makes any warranty, express or implied, or assumes any legal liability or responsibility for the accuracy, completeness, or usefulness of any information, apparatus, product, or process disclosed, or represents that its use would not infringe privately owned rights. Reference herein to any specific commercial product, process, or service by trade name, trademark, manufacturer, or otherwise does not necessarily constitute or imply its endorsement, recommendation, or favoring by the United States Government or any agency thereof. The views and opinions of authors expressed herein do not necessarily state or reflect those of the United States Government or any agency thereof.

\section*{Data Availability}
The synthetic data that supports the findings of this study are available within the article. The weekly see surface temperature data can be found at the NOAA Optimum Interpolation (OI) SST V2 webpage: \url{https://psl.noaa.gov/data/gridded/data.noaa.oisst.v2.html}. The datasets used and/or analysed during the current study are also available from the corresponding author on reasonable request.

%\section{Bibliography styles}
%\section*{References}
\bibliographystyle{elsarticle-harv}
\bibliography{references}
\end{document}